\def\year{2021}\relax
\pdfoutput=1
\documentclass[letterpaper]{article} 
\usepackage{aaai21}  
\usepackage{times}  
\usepackage{helvet} 
\usepackage{courier}  
\usepackage[hyphens]{url}  
\usepackage{graphicx} 
\usepackage{natbib}  
\usepackage{caption} 
\usepackage{amsmath}
\usepackage{amsfonts}
\usepackage{makecell}
\usepackage{amssymb}

\title{Exploiting Behavioral Consistence for Universal User Representation }
\author{
    Jie Gu\thanks{Equal contribution.},
    Feng Wang$^{*}$,
    Qinghui Sun, \\
    Zhiquan Ye,
    Xiaoxiao Xu,
    Jingmin Chen,
    Jun Zhang \\
}
\affiliations{

    Alibaba Group, Hangzhou, China \\
    \{yemu.gj,wf135777,yuyang.sqh,beichen.yzq,xiaoxiao.xuxx,jingmin.cjm,zj157077\}@alibaba-inc.com

}

\begin{document}

\maketitle

\begin{abstract}
User modeling is critical for developing personalized services in industry. A common way for user modeling is to learn user representations that can be distinguished by their interests or preferences. In this work, we focus on developing universal user representation model. The obtained universal representations are expected to contain rich information, and be applicable to various downstream applications without further modifications (\emph{e.g.}, user preference prediction and user profiling). Accordingly, we can be free from the heavy work of training task-specific models for every downstream task as in previous works. In specific, we propose Self-supervised User Modeling Network (SUMN) to encode behavior data into the universal representation. It includes two key components. The first one is a new learning objective, which guides the model to fully identify and preserve valuable user information under a self-supervised learning framework. The other one is a multi-hop aggregation layer, which benefits the model capacity in aggregating diverse behaviors.
Extensive experiments on benchmark datasets show that our approach can outperform state-of-the-art unsupervised representation methods, and even compete with supervised ones.
\end{abstract}

\section{Introduction} \label{sec:intro}
User modeling is one of the most important issues in the era of big data. It helps business to better understand behaviors of users and then improve the quality of commercial services. A recent common solution is to learn an encoder to convert collected user behavior data (\emph{e.g.}, search queries in e-commerce platform) into low dimensional representations with neural networks. Such an end-to-end learning paradigm can not only reduce feature engineering work but also improve performance compared to hand-crafted methods. 

A large proportion of current user modeling approaches are task-specific. That is, they are trained specifically for certain downstream applications in a supervised fashion, \emph{e.g.}, for click rate prediction and personalized recommendation \cite{zhou2018deep,yu2019adaptive,li2019multi}. Since the models would be easily biased to the learning task, the generalization ability of these task-specific user representations can probably be limited. They might only perform well in the designed application scenarios but fail on other unseen tasks. Accordingly, to ensure performance, we have to train a particular representation model for each downstream task. But this is time-consuming and would cost a lot of computing and storage resources.

In contrast, this work focuses on learning universal user representations.  Basically, we attempt to learn a behavior data encoder on a large amount of unlabeled user logs to infer user representations. The obtained representations are expected to be informative and applicable to various unseen downstream tasks (like preference prediction).
Compared to the task-specific ones, universal representation model shows great advantages in reducing engineering works, saving time and computational costs.
Despite the potential advantages, so far there are only few public works discussing this subject \cite{benton2016learning,ding2017multi,robertson2004understanding,andrews2019learning}.
One challenge in learning such a universal encoder is to design a proper task to guide the encoder being capable of fully identifying valuable user information.

Through data exploration and analysis, we observe that most of our users exhibit consistent behaviors over a relatively long time span. This is probably because  behaviors of users are usually predominated by some consistent or slowly evolved factors like gender, age, hobbies, etc. For example, a man may have long-term and consistent interests in searching men's clothing/shoes. Such a kind of behavioral consistency motivates us to develop a new learning task by mining the implicit dependencies between behaviors.

Accordingly, a novel objective named as behavioral consistency loss is presented. During training, each sample consists of two disjoint sets of user behaviors, \emph{i.e.}, the input and target sets respectively. 
The encoder generates the universal user representation based on the input behavior data. Then the loss is built by predicting the number of occurrences of every element in the target set.
Such an objective guides the representation model to fully discover and characterize the underlying consistent factors shared within user behaviors. The learned universal representations would thus retain rich information, leading to a better performance and wider range of applications. Our method falls into the self-supervised learning paradigm in the sense that user behavior sequence itself can provide supervisory signal \cite{lecunssl2020aaai}. 

Besides the need of a proper learning task, another key to the success of achieving satisfactory user representations is the design of neural network. In the literature, most existing works follow the same design paradigm: first encoding variable-sized behavior items (\emph{e.g.}, posted tweets or search queries) as embeddings and then applying some kind of information aggregation module to obtain the fixed-sized user representation. A common way for aggregation is to use max (or average) pooling or attention mechanism. However, in our case, the user behavior sequences are usually not semantically coherent as in NLP applications like sentence embedding. Therefore, we argue that simple pooling or attention can hardly characterize users well due to the diversity of behaviors. To alleviate this issue, we propose a multi-hop aggregation layer to increase the model capacity in summarizing diverse aspects of users. It is realized by performing attention multiple times with shared parameters, refining user representation iteratively. 

The framework is named as Self-supervised User Modeling Network, termed as SUMN. The main contributions of this paper are as follows:

\begin{itemize}
\item Universal user modeling is an important but less-explored topic. In this work, we propose to train effective universal user representations with a novel objective named behavioral consistency loss. The learned representations are informative and can be directly applied to various downstream tasks, \emph{e.g.}, predicting user preferences and inferring user profiles.

\item We identify the limitation of using simple pooling to~summarize user behavior embeddings and introduce a multi-hop aggregation layer. This layer employs multi-hop attention mechanism to incrementally refine user representation. It can improve the model capacity and better tackle the diversity dilemma.

\item We conduct extensive experiments on three datasets collected from e-commerce platforms and public social media websites. The results in these experiments demonstrate the effectiveness and transferability of the learned user representations. 
\end{itemize}

To the best of our knowledge, this work is one of the pioneering works showing that it is possible to obtain generic user representations via self-supervised learning. Moreover, the obtained user representations can be seamlessly applied to a board range of downstream applications and achieve promising results. 
Source codes of SUMN will be released along with the final version.

\section{Related Work} \label{sec:relwork}
\textbf{Task-specific User Representation Learning} There is a rich literature on task-specific user representation modeling \cite{yu2019adaptive,li2019multi,an2019neural}. These models are typically learned in a supervised manner for one or several particular applications. To achieve better performance on downstream applications, the model components for representation extraction and target task mapping are usually trained simultaneously \cite{xue2017deep,yu2019adaptive,ni2018perceive,li2019multi}. But with such an end-to-end learning strategy, the information unrelated to the target task is quite likely to be ignored and discarded. Thus the generalization abilities of these specifically learned user representations are probably limited, which means that they can hardly be applied or transferred to other unseen scenarios directly. Moreover, some task-specific user modeling methods \cite{chen2018sequential,an2019neural} require additional rounds of gradient descent to obtain the representations for new users. They might achieve better applicability, but are more time-consuming and less efficient.

\textbf{Universal User Representation Learning} Compared to the task-specific user modeling, the study about the universal user representation learning is still in  early stage. Existing methods generally adopt dimensionality reduction (\emph{e.g.}, PCA and autoencoder) or word/document embedding techniques to generate user representations \cite{yu2016user,amir2016modelling,benton2016learning,ding2017multi}. Beyond that, \cite{ni2018perceive} proposes to learn universal user embeddings with multiple supervised tasks. Despite the improvements, the generalization ability of this method may still suffer due to the requirement of annotated training set and the lack of principles of selecting proper training tasks. Recently, \cite{andrews2019learning} explores to achieve satisfactory user representations with an identity prediction based objective. It focuses on mining distinctive features for predicting user's identity. We found that such a~representation model is more suitable for dealing with matching-related tasks, while less effective on other downstream tasks like user profiling.

\textbf{Feature Pooling} Average and max pooling are the most widely used modules for feature aggregation in neural networks. To realize a learnable and more effective pooling, one typical alternative network module in the NLP field is attention \cite{yang2016hierarchical}. The main idea is to learn a fixed~context vector to measure the importance of each feature and then aggregate them by weighted summation. However, we argue that using only a single attention module can hardly characterize users well due to the diversity of behaviors. In contrast, multi-hop attention \cite{sukhbaatar2015end,miller2016key} (can be viewed as a form of memory) is capable of capturing rich information from different aspects of inputs, refining user representations iteratively. Such an advantage of multi-hop attention exactly corresponds with our demands for the feature pooling strategy.

\section{Learning Universal Representation} \label{sec:model}
\begin{figure*}[t]
    \centering
    \includegraphics[width=0.8\textwidth]{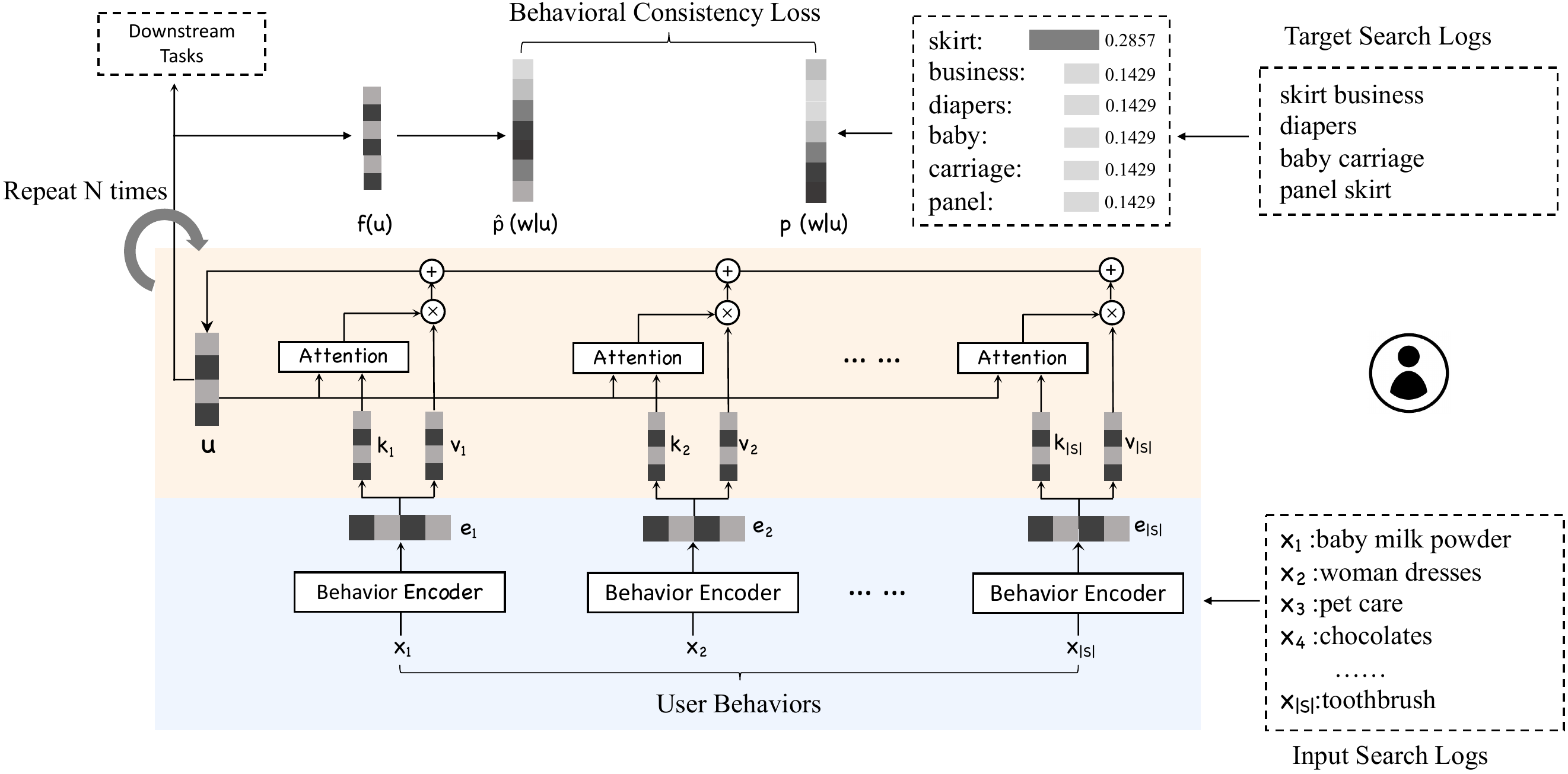}
    \caption{Illustration of our SUMN approach for learning universal user representations with search logs as inputs. The representation encoder is trained once. The inferred user representations can be applied to various downstream tasks, like preference identification and user profiling (with only further training simple corresponding classifiers).}
    \label{figure:overview}
\end{figure*}

In this work, we focus on mining \emph{long-term} interests of users for addressing various business needs in e-commerce platforms or social networks, \emph{e.g.}, user profiling and preference prediction. Our solution for this issue is to achieve an universal user representation encoder. For example, a man who loves outside activities may probably have many search logs about men’s outdoor goods. Mining such long-term preferences and user properties to produce effective universal user representation is the subject of this work. Once the universal representations are obtained, they can be directly applied to  downstream tasks without further fine-tuning. We can be free from the heavy work of training task-specific representation models as in previous works.

For this purpose, we propose a Self-supervised User Modeling Network (SUMN) to encode raw historical behaviors $S$ into user representation $\mathbf{u} \in \mathbb{R}^d$ ($d$ is the embedding dimension). The overview of SUMN is shown in Figure \ref{figure:overview}. 
Generally, SUMN is trained in a self-supervised manner with a novel objective named behavioral consistency loss. This is motivated by the observation that there are some consistent (or slowly changed) underlying features that predominate one's behaviors over a long time (\emph{e.g.}, age, gender, habits). Moreover, a multi-hop aggregation layer is further introduced, which refines user representations via multi-pass attention. Such a module helps increase the model capacity in summarizing diverse aspects of user behaviors.

\subsection{Preliminaries}
Let $U = \{u_1, u_2, \cdots, u_n\}$ be a set of users. Given a user, his/her behavior history can be formulized as a set, \emph{i.e.}, $S=\{x_1, \dots, x_{|S|}\}$. Specifically, $x_t$ denotes the context or content of the $t$-th behavior, and $|S|$ is the total number of the historical behaviors. The goal of user representation learning is to achieve an encoder that is capable of generating low-dimensional yet informative representations based on historical behaviors. 

As shown in Figure \ref{figure:overview}, there are three main components in SUMN. The first one is a behavior encoder, which learns to transform the behaviors in $S$ into corresponding embeddings, \emph{i.e.},

\begin{equation}
\mathbf{e}_t=f_{behavior}(x_t),
\end{equation}
where $\mathbf{e}_t\in\mathbb{R}^d$ denotes the embedding of $x_t$, with dimension of $d$. 

Since unstructured text is ubiquitous in user logs and usually contains rich information, without loss of generality, we focus on the text modality in this work. Accordingly, the contents of historical behaviors are supposed to be their own texts (\emph{e.g.}, the visiting behavior can be naturally formed as the title of the visited good). Formally, $x_t$  can be further expressed as $[w_1^t, \cdots, w_{|x_t|}^t]$, where $w_i^t$ denotes a word drawn from a vocabulary $\mathcal{V}$, and $|x_t|$ denotes the length of $x_t$ (namely the number of words). The behavior encoder $f_{behavior}$ turns out to be a function for text modeling. Here we adopt mean word embedding, namely $f_{behavior}(x)=\frac{1}{|x|}\sum_i^{|x|}\mathbf{w}_i$, where $\mathbf{w}_i$ is the embedding of the word $w_i$. Compared to more recent text embedding 
methods, \emph{e.g.}, BERT \cite{devlin2019bert}, mean embedding has a great advantage in efficiency. Since one person usually have hundreds of behaviors in real-world applications, it is time-consuming and computationally expensive to encode behaviors with BERT.

The variable-sized behavior embeddings are then sent to a new aggregation layer to yield the fixed-sized user representation, \emph{i.e.},

\begin{equation}
\label{eq:aggregate}
\mathbf{u}=\mathrm{Aggregate}(\{\mathbf{e}_1, \cdots, \mathbf{e}_{|S|}\}),
\end{equation}
where $\mathbf{u}\in\mathbb{R}^d$ denotes the user representation. The aggregation layer uses multi-hop attention and can better retain the diversity of behaviors compared to simple pooling approaches. More details are given in the ``Multi-hop Aggregation Layer'' section. 

The third key component is the behavioral consistency loss. It takes $\mathbf{u}$ as inputs and guides the model to predict the distribution of the word occurrences in a target behavior set $T=\{y_1, \dots, y_{|T|}\}$. The historical and target behavior sets are disjoint, and $y_t$ has the same definition as $x$. The learning process asks SUMN to discover implicit dependencies between user behaviors. Refer to the following section for more details.

\subsection{Behavioral Consistency Loss}
With behavioral consistency loss, SUMN is driven to yield generic and meaningful user representations. Specifically, a user representation $\mathbf{u}$ can be derived after aggregating the behavior embeddings. Based on $\mathbf{u}$, the model is then asked to predict the occurrence of every element (\emph{i.e.}, word) in a target behavior set $T$. The target set is formed with the behaviors that are not overlapped with historical ones in terms of time. The optimization is performed by minimizing the gap between the ground-truth and predictions.

The behavioral consistency loss is motivated by the observation that behaviors of users are usually affected by some consistent or slowly changed latent factors, \emph{e.g.}, age, gender, hobbies (reflecting long-term behavioral habits). Thus there exists dependencies/relations between behaviors in different time periods. By predicting some statistics of target behaviors, the user representation encoder is driven to well characterize the latent consistent factors from historical behaviors. Accordingly, the~generated user representation would be informative and interest- or preference-aware. 

In practice, all elements in $T$ are included in~the loss function to preserve more information regardless of their popularity. The elements refer to the words appeared in $y_t$ $(t=1,\dots,|T|)$ since we have restricted the contents of behaviors to be text. For simplicity, we collect ``future" behaviors happened after historical ones to build the set $T$ in this work (though other alternatives might also work). With user embedding $\mathbf{u}$ as input, the distribution of word occurrences can be predicted as 

\begin{equation}
\hat{p}(w|u)=\frac{\exp(\mathbf{o}_w^Tf(\mathbf{u}))}{\sum_{w'\in\mathcal{V}} \exp(\mathbf{o}_{w'}^Tf(\mathbf{u}))},
\end{equation}
where $f(\mathbf{u})$ is a fully connected layer, \emph{i.e.}, $f(\mathbf{u})=\mathrm{ReLU}({\mathbf{W}^o\mathbf{u}})$ ($\mathbf{W}^o\in\mathbb{R}^{d\times d}$ is the parameter matrix for linear transformation). $w, w'\in\mathcal{V}$ denotes a word in target behaviors, and the vector $\mathbf{o}_w \in\mathbb{R}^d$ is the mapping parameter corresponding to the word $w$ for the occurrence prediction.

The ground-truth is the normalized occurrences of words, \emph{i.e.},
\begin{equation}
p(w|u) = \frac{\log(1+\mathrm{count}(w))}{\sum_{w'\in\mathcal{V}}\log(1+\mathrm{count(w)})},
\end{equation}
where $\mathrm{count}(w)$ is the number of occurrence of the word $w$ in $T$. An example is given in Figure~\ref{figure:overview}. The log operation could prevent the parts of common words from occupying the most of the distribution. Given $p(w|u)$ and $\hat{p}(w|u)$, the Kullback–Leibler divergence is used to measure the differences, which defines the loss function:
\begin{equation}
\label{eq:loss_func}
\mathcal{L}=D_{KL}(p||\hat{p})\propto-\sum_{w\in\mathcal{V}} p(w|u)\log\hat{p}(w|u).
\end{equation}

\subsection{Multi-hop Aggregation Layer}
An aggregation layer is required to convert a variable number of behavior embeddings $\{\mathbf{e}_1,\mathbf{e}_2,\cdots,$ $\mathbf{e}_{|S|}\}$ into a fixed-sized user representation $\mathbf{u}$ (Equation~(\ref{eq:aggregate})). In SUMN, we introduce a new multi-hop aggregation module. The architecture is a form of memory layer \cite{sukhbaatar2015end}.

In real-world applications, we found that such an aggregation achieves a desirable trade-off between efficiency and effectiveness in information integration. It is capable of mining diverse aspects of users compared to simple max or average pooling operations. Furthermore, in comparison with more complicated architectures (\emph{e.g.}, self-attention based aggregation), multi-hop aggregation mechanism excels at efficiency, especially when dealing with hundreds of historical behavior embeddings.

Specifically, suppose there are $N\in\mathbb{N_+}$ hops. In every hop, the behavior embeddings are projected to their own key and value embeddings with a linear transformation, \emph{i.e.},
\begin{equation}
\label{eq:keys}
\mathbf{k}_t=\mathrm{LN}(\mathbf{W}^K\mathbf{e}_t) \quad
\mathbf{v}_t=\mathrm{LN}(\mathbf{W}^V\mathbf{e}_t),
\end{equation}
where $\mathbf{W}^K\in\mathbb{R}^{d \times d}$ and $\mathbf{W}^V\in\mathbb{R}^{d\times d}$ are parameter matrices for transformation, which are shared in each hop. $\mathrm{LN}$ stands for the layer normalization layer \cite{ba2016layer}.

For each user, let $\mathbf{u}^{(h-1)}$ be the refined user representation after $h-1$ hops. The attention scores used in the $h$-th hop can be computed as 
\begin{equation}
\alpha^{(h)}_t=\frac{\exp(\mathbf{k}_t^T\mathbf{u}^{(h-1)})}{\sum_{j=1}^{|S|}\exp(\mathbf{k}_j^T\mathbf{u}^{(h-1)})}.
\end{equation}

The aggregation layer maintains a memory vector $\mathbf{m}^{(h)}$ to accumulate collected information after the first $h$ hop. The attention scores indicate which behavior should be emphasized or neglected according to current collected information. Subsequently, the memory update $\Delta \mathbf{m}^{(h)}$ in the $h$-th hop is constructed by the weighted sum of the value embeddings, namely $\Delta \mathbf{m}^{(h)}=\sum_{t=1}^{|S|} \alpha^{(h)}_t \mathbf{v}_t$. Then the update is applied to the current memory vector as $\mathbf{m}^{(h)} = \mathbf{m}^{(h-1)} + \Delta \mathbf{m}^{(h)}$.

The user representation is the layer-normalized memory vector for consistent numerical scale:
\begin{equation}
\mathbf{u}^{(h)} = \mathrm{LN}(\mathbf{m}^{(h)}).
\end{equation}
The initial memory vector, denoted as $\mathbf{m}^{(0)}$, is randomly initialized and needed to be trained with error back propagation. We further set $\mathbf{u}^{(0)}=\mathbf{m}^{(0)}$.

The multi-hop aggregation layer outputs $N$ user representations $[\mathbf{u}^{(1)}, \cdots,\mathbf{u}^{(N)}]$, corresponding to $N$ hops respectively. In training phase, we perform training with the same loss on all these representations to alleviate long-term memorization burden. During testing, only $\mathbf{u}^{(N)}$ is kept as the final user embedding while the others are discarded.

\section{Experiments} \label{sec:experiments}

In this section, we report extensive experimental~results on multiple real-world datasets and tasks. The thorough comparison with several state-of-the-art methods verifies the superiority of our SUMN.

\subsection{Datasets $\&$ Implementation Details}
The experiments are conducted on two e-commerce datasets and one social media dataset. Each dataset consists of large-scale unlabelled user behaviors for learning meaningful user representations. The types of text include product descriptions, tweets and search queries. An overview of the datasets are summarized in Table \ref{tab:data_stat}.

\textbf{Amazon Dataset\footnote{https://nijianmo.github.io/amazon/index.html}} This dataset includes product reviews and involved product metadata like titles and categories. For each user, the reviewed product titles make up a review behavior sequence. We~selected the review logs happened between 2014-07 and 2014-12 to form the historical set $S$, while taking the review logs happened between 2015-01 and 2015-06 for the target set $T$. Users without review logs in either of these two periods are omitted. 

\textbf{Twitter Dataset} We download the twitter archives between 2016-08 and 2017-01 from the Internet Archive \footnote{https://archive.org/} to evaluate the performance of our SUMN on social media data. For each person, his/her posted tweets are collected to build the history/target behavior set. Specifically, all re-tweets and non-English tweets are filtered out. The tweets posted from 2016-08 to 2016-10 are chosen for $S$, while the rest ones are selected for $T$.

\begin{table}
\centering
\begin{tabular}{lcp{0.8cm}<{\centering}p{0.8cm}<{\centering}c}
\hline
Dataset  & $|U|$ & UL.$|S|$ &UL.$|x|$ & $|\mathcal{V}|$ \\
\hline
Amazon & {\footnotesize 1,725,907} & {\footnotesize 25} & {\footnotesize 35} & {\footnotesize 50,000}  \\
Twitter & {\footnotesize 741,279} & {\footnotesize 10} & {\footnotesize 12} & {\footnotesize 30,000}      \\
Industrial & {\footnotesize 64,000,000} & {\footnotesize 320} & {\footnotesize 8} & {\footnotesize 178,422}     \\
\hline
\end{tabular}
\caption{Statistics of the datasets.\; The symbol UL.~indicates the truncation threshold.}
\label{tab:data_stat}
\end{table}

\textbf{Industrial Dataset} We also build a dataset by collecting search logs on a popular e-commerce platform to verify the effectiveness of SUMN in real-world scenarios. We randomly sampled 64~million users who have search logs both in 2019-02$\thicksim$ 2019-03 and 2019-06$\thicksim$2019-07. Queries in the former period are collected for building $S$, and those in the latter period are chosen for $T$.

\textbf{Data Prepossessing} For English texts, we perform the operations of lowercasing and word stemming. The Chinese texts are segmented by using Jieba\footnote{https://github.com/fxsjy/jieba}. A dedicated vocabulary is constructed for each dataset. We also set truncation thresholds to limit the number of behaviors in $S$, as well as the number of words in $x$. The exceeded behaviors or words are removed. The principle of the setting of truncation threshold is that $95\%$ data values can be covered by the threshold (\emph{e.g.}, in Amazon Dataset, the threshold for $|S|$ is set as 25 since $95\%$ of the size of the historical sets are smaller than 25). The statistics of vocabularies and truncation thresholds are listed in Table \ref{tab:data_stat}.

\textbf{Parameter Setting of SUMN} For all datasets, the dimension of all embeddings in SUMN, namely $d$, is set to be 256, and the number of hops is set to be 5. The loss function (Equation \ref{eq:loss_func}) is optimized by the Adam optimizer \cite{kingma2014adam} with a learning rate of $0.001$ and a batch size of $256$. The training is stopped when the loss converges on the validation set. 

\subsection{Evaluations on Downstream Tasks}
We compare the performance of our approach against other representative methods on two downstream tasks. Specifically, one is category preference identification and the other is user profiling.

\begin{table}
\newcommand{\tabincell}[2]{\begin{tabular}{@{}#1@{}}#2\end{tabular}}
\centering
\begin{tabular}{p{2.4cm}p{1.3cm}<{\centering}p{1.2cm}<{\centering}p{1.25cm}<{\centering}}
\hline
Method  & {\footnotesize \tabincell{c}{ Computers \& \\ Technology}} & \footnotesize \tabincell{c}{ Video \\ Games} & \footnotesize \tabincell{c}{ Costumes \& \\ Accessories}  \\
\hline
TextCNN & 0.7768 & 0.7953  & 0.7747 \\
HAN & \textbf{0.8092} & \textbf{0.8235}  & \textbf{0.8072} \\
\hline
TFIDF & 0.6733 & 0.7360  & 0.7102\\
Word2Vec& 0.7526 & 0.7805  & 0.7622\\
Doc2Vec & 0.7658 & 0.7917  & 0.7775 \\
\hline
SUMN-AE & 0.7662 & 0.7979  & 0.7753 \\
SUMN-ID & 0.7771 & 0.8021  & 0.7832 \\
SUMN-MEAN & 0.7569 & 0.7893  & 0.7670 \\
SUMN-MAX & 0.7851 & 0.8038  & 0.7814 \\
\hline
SUMN & \textbf{0.7948} & \textbf{0.8245}  & \textbf{0.8010}  \\
\hline
\end{tabular}
\caption{Performance comparison in terms of AUC on Amazon dataset. The downstream task is to predict category preferences. We use \textbf{bold} font to highlight wins.}
\label{tab:amazon_result}
\end{table}

\textbf{Data Collection} Category preference identification refers to the task of predicting whether a user have a long-term preference for the commodities in the target category. This requires to capture user's shopping interests. 

For this task, we conduct experiments on two e-commerce datasets. For Amazon dataset, three categories are included: books of computers $\&$~technology, video games and costumes $\&$ accessories. Review logs between 2015-07 and 2015-12 are collected to infer the user representations, and a user is labeled as positive if there exists at least one review log between 2016-07 and 2016-12. A total of 191,856 samples are collected. For the industrial dataset, we consider three categories including outdoor products, children's products and car accessories for evaluation. We collect search queries from 2019-05 to 2019-06 for user representation inference, and transaction logs between 2019-08 and 2019-09 for user labeling, which makes up a set containing 2.4 million samples. For all evaluation datasets, we randomly select 80\% of the samples for training downstream models and the rest for performance test.

User profiling prediction aims to identify user aspects such as gender and age. We also conduct experiments on two datasets. For Twitter dataset, the evaluation task is gender classification. The~evaluation dataset is the Author Profiling dataset\footnote{https://pan.webis.de/clef17/pan17-web/author-profiling.html}, which provides gender labels of twitter users, as well as corresponding tweet logs. For each user, we sample at most ten tweets as inputs to SUMN to generate user representations. There are 3,000 samples collected for training and 1,900 samples for test. For industrial dataset, we evaluate performances on two sub-tasks: (1) user age classification task (6-class), which predicts the age ranges of users. There are 1,628,958 samples for training and 543,561 samples for testing; (2) baby age classification task (7-class), which predicts the age ranges of users' babies. The sizes of the training and testing sets are 396,749 and 99,411 respectively. Search queries are collected for user representation inference. The ground-truth age label comes from an online questionnaire.

\begin{table}
\newcommand{\tabincell}[2]{\begin{tabular}{@{}#1@{}}#2\end{tabular}}
\centering
\begin{tabular}{p{2.4cm}p{1.2cm}<{\centering}p{1.2cm}<{\centering}p{1.3cm}<{\centering}}
\hline
Method  & {\footnotesize\tabincell{c}{Outdoor\\ Products}} & {\footnotesize \tabincell{c}{Children's\\Products}} & {\footnotesize  \tabincell{c}{Car \\ Accessories}}  \\
\hline
TextCNN & 0.7627 & 0.8538  & 0.8521 \\
\hline
TFIDF & 0.6720 & 0.7738  & 0.8118\\
Word2Vec& 0.7032 & 0.7982  & 0.8203\\
Doc2Vec & 0.7552 & 0.8381  & 0.8443 \\
\hline
SUMN-AE & 0.7583 & 0.8550  & 0.8525 \\
SUMN-ID & 0.7576 & 0.8503  & 0.8397 \\
SUMN-MEAN & 0.7102 & 0.8112  & 0.8363 \\
SUMN-MAX & 0.7461 & 0.8416  & 0.8431 \\
\hline
SUMN & \textbf{0.7784} & \textbf{0.8700}  & \textbf{0.8701}  \\
\hline
\end{tabular}
\caption{Performance comparison in terms of AUC on industrial dataset. The downstream task is to predict category preferences. We use \textbf{bold} font to show wins.}
\label{tab:industrial_result}
\end{table}

\textbf{Competitors}
Two types of previous representative approaches are selected for comparison. The first class of methods generate (unsupervised) user representations without access to the annotated data labels in downstream tasks. The competitors include: (1)\textit{TF-IDF} \cite{robertson2004understanding}, which views texts in one's behaviors as a single document and uses a sparse statistical vector for user representation; (2)\textit{Word2Vec} \cite{mikolov2013distributed}, training word embeddings on an unlabelled corpus and computing user representations as the average of the word embeddings in behaviors; (3)\textit{Doc2Vec} \cite{le2014distributed}, regarding the behaviors of a person as a document and learning a document embedding for user representation. The other type of competitors simultaneously learn task-specific representation encoders and classifiers in a supervised manner on downstream tasks. Specifically, they are: (1)\textit{TextCNN} \cite{kim2014convolutional}, applying convolution operations on the embedding concatenation of all words appeared in behaviors and using max pooling to get user representations; (2)\textit{HAN} \cite{yang2016hierarchical}, employing a hierarchical attention network, where two levels of attention operations are adopted to aggregate word and behavior embeddings respectively; (3)\textit{BERT} \cite{devlin2019bert}, training a deep language model on a large corpus and being finetuned on downstream tasks. In our case, the model input is the concatenated behavior texts separated by a special token `[SEP]'.

\textbf{Downstream Models} For unsupervised methods, the downstream models are implemented by MLP classifiers applied after the derived representations (generated by SUMN or competitors). The MLP has only one single hidden layer with the dimension set as 128. Moreover, we use pre-trained word embeddings to initialize TextCNN and HAN to prevent overfitting. The hyper-parameters of the supervised competitors are tuned on the validation set. For both the MLP and supervised models, we use Adam with a learning rate of 0.001 as the~optimizer, and the batch size is set as 256.

Tables \ref{tab:amazon_result} and \ref{tab:industrial_result} show the results of category~preference identification on the Amazon and industrial datasets, respectively. Tables \ref{tab:profile_result} and \ref{tab:profile_result2} list the comparisons on the two datasets of user profiling prediction. Results are reported by ourselves\footnote{The results of HAN and BERT are not available in some cases because the model training is extremely slow and costs too many computing resources (there are hundreds or even thousands of words as model inputs) in these cases.}. We have two observations. First, SUMN consistently outperforms the unsupervised competitors, \emph{e.g.}, about $0.027$ average AUC improvements than Doc2Vec on the industrial dataset in predicting category preferences. Second, our model achieves comparable performance to supervised methods, \emph{e.g.}, about $1\%$ higher accuracy than TextCNN and HAN on the tweet dataset. It is noteworthy that the aim of this work is not to develop highly effective task-specific representation encoders. The cost would be high if we always train a supervised model for a new task. In contrast, SUMN generates universal user representations that are shared for downstream tasks. The universal representation can be easily applied to various tasks with only training a simple MLP, which meets the requirements of efficiency, easy-using and effectiveness in real-world applications.

\begin{table}
\centering
\begin{tabular}{lccc}
\hline
Method  & Accuracy & AUC  \\
\hline
TextCNN & 69.15 & 0.7538 \\
HAN & 69.21 & 0.7633 \\
BERT  & \textbf{70.79} & \textbf{0.7767} \\
\hline

TFIDF & 67.54 & 0.7259 \\
Word2Vec& 67.98 & 0.7368 \\
Doc2Vec & 68.42 & 0.7451 \\
\hline
SUMN-AE  & 68.54 & 0.7517  \\
SUMN-ID & 67.16 & 0.7398  \\
SUMN-MEAN & 69.33  & 0.7594 \\
SUMN-MAX & 67.94 & 0.7326\\
\hline
SUMN & \textbf{70.28} & \textbf{0.7728}  \\
\hline
\end{tabular}
\caption{Performance comparison on gender prediction of twitter users in terms of accuracy and AUC. We use \textbf{bold} font to highlight wins.}
\label{tab:profile_result}
\end{table}

\begin{table}
\centering
\begin{tabular}{lccc}
\hline
Method  & Age & Baby Age  \\
\hline
TextCNN & 60.64 & 72.03 \\
\hline

TFIDF & 61.22 & 69.06 \\
Word2Vec& 57.52 & 66.66 \\
Doc2Vec & 53.51 & 65.86 \\
\hline
SUMN-AE  & 61.93 & 67.30  \\
SUMN-ID & 56.62 & 64.85  \\
SUMN-MEAN & 62.14  & 70.84 \\
SUMN-MAX & 61.55 & 70.03\\
\hline
SUMN & \textbf{64.47} & \textbf{72.17}  \\
\hline
\end{tabular}
\caption{Performance comparison on age and baby-age predictions in terms of accuracy on industrial dataset. We use \textbf{bold} font to highlight wins.}
\label{tab:profile_result2}
\end{table}

\subsection{Ablation Study $\&$ Discussion}
In this part, we aim to show that the performance of SUMN benefits from the introduced components. Moreover, two performance issues are discussed.

\textbf{Ablation Study}
We use 4 variants of SUMN as baselines for comparison: (1)\textit{SUMN-AE}, where the historical and target behavior sets are set to be the same, namely $T=S$; (2)\textit{SUMN-ID}, replacing the objective in SUMN with a identity prediction task (also designed for user representation learning presented by \cite{andrews2019learning}); (3) \textit{SUMN-MEAN}, using mean pooling to replace the multi-hop aggregation layer; (4)\textit{SUMN-MAX}, using max pooling for aggregation. All training configurations and parameters are set to be the same as those of SUMN.

One can observe from tables~\ref{tab:amazon_result}-\ref{tab:profile_result2} that SUMN works consistently better than SUMN-(AE,ID) and SUMN-(MEAN,MAX). The comparison demonstrates the effectiveness of the presented behavioral consistency loss and multi-hop aggregation layer. 

\begin{figure}[t]
    \centering
    \includegraphics[width=0.43\textwidth]{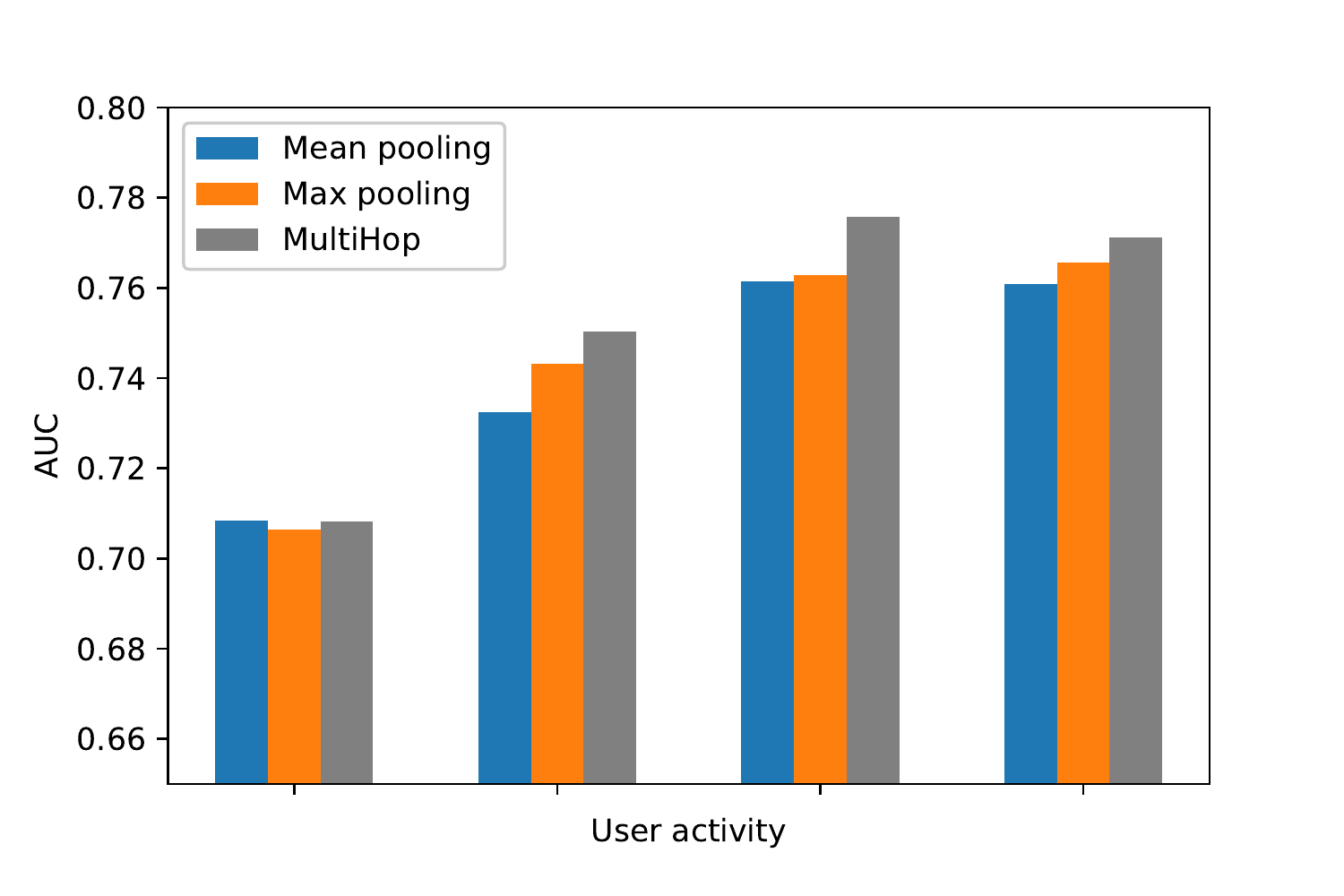}
    \caption{Performance comparison of applying different aggregation mechanisms to different user sets (having different numbers of activities).}
    \label{fig:seq_multihop}
\end{figure}

\textbf{Effects of Multi-hop Aggregation}
The experiment is conducted on the Amazon evaluation~data-set prepared for the preference prediction task of video-game category. Specifically, we divide users into four disjoint sets according to the numbers~of their activities. The representations of these users in four sets are generated by SUMN, SUMN-MAX and SUMN-MEAN, respectively. The comparisons are shown in Figure \ref{fig:seq_multihop}, in which the results are reported on hold-out validation sets. One can observe that multi-hop aggregation can better characterize users with more (usually diverse) behaviors than max or average pooling.

\begin{figure}[t]
    \centering
    \includegraphics[width=0.43\textwidth]{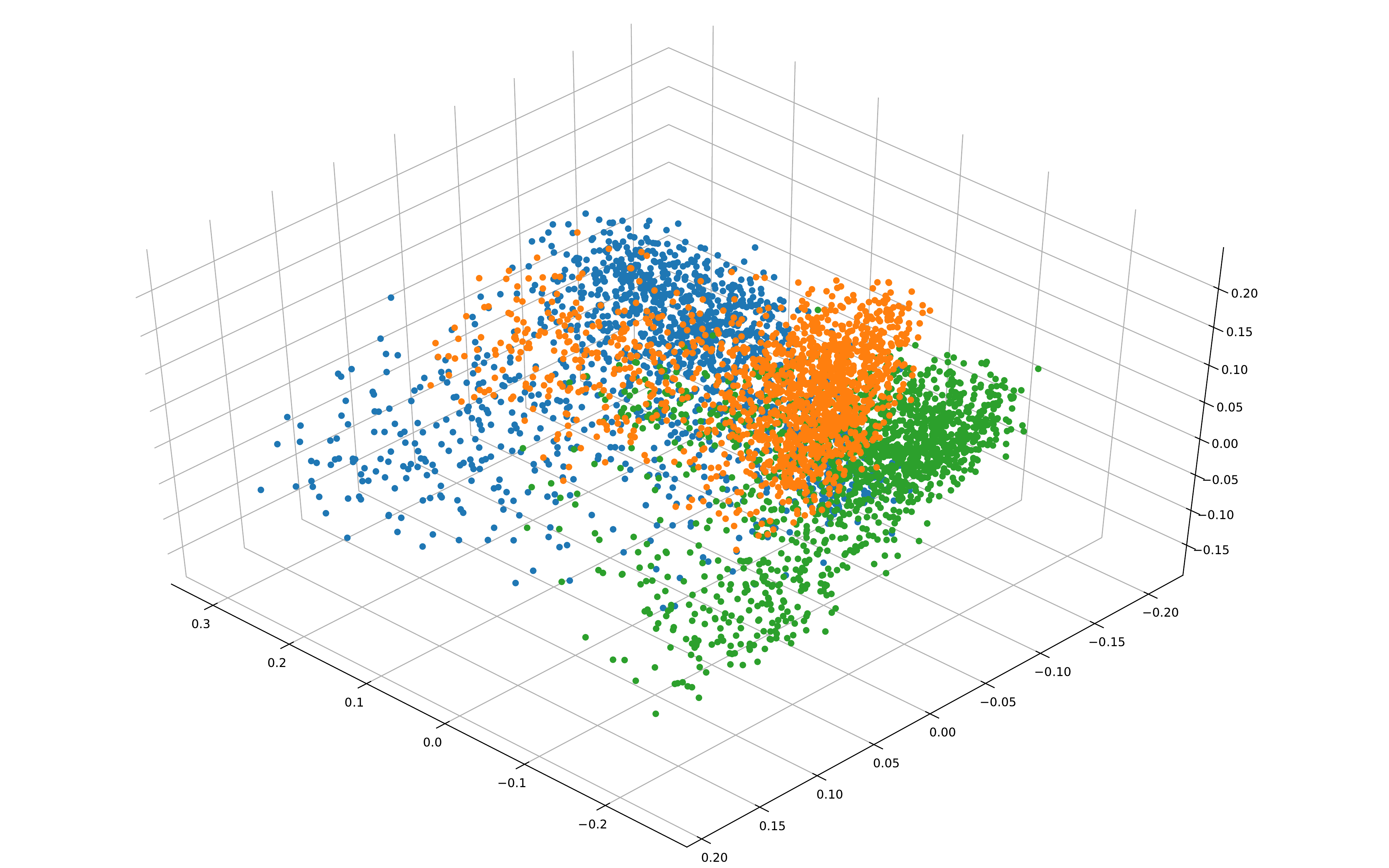}
    \caption{Visualization of the first three principal components of user representations. The dots with the~same color refer to the embeddings of users with the same category preference.}
    \label{fig:embedding_viz}
\end{figure}

\textbf{Visualization of Representation}
We generate representations for users in the industrial evaluation dataset prepared for the category preference prediction experiments. PCA \cite{rokhlin2010randomized} is then applied to the derived user representations. Figure \ref{fig:embedding_viz} visualizes the first three principal components, where only users with positive labels are included. We can see that the users with the same category preference are embedded more close, though the representation model SUMN is trained with no access to preference labels.

\textbf{Discussion}
SUMN does not explicitly encode sequence information. One may wonder whether the performance can be further improved if we replace the multi-hop aggregation layer with a sequence data encoder, \emph{e.g.}, transformer. The experimental results show that though the transformer module introduces more parameters (which is computationally expensive and memory intensive), the performance improvement is very marginal (less than 1$\%$ on the task of user age prediction). This is predictable since long-term preference and user properties usually evolve slowly. Thus capturing such evolving has minor effects on the performance.


\section{Conclusion} \label{sec:conclusion}
In this paper, we propose a self-supervised approach (SUMN) for universal user representation learning. Inspired by the observation that the user behaviors show some inherent consistency, we design a novel learning objective named behavioral consistency loss to guide the model to extract latent user factors. To further tackle the diversity problem, we propose a multi-hop aggregation layer which leverages multi-hop attention to iteratively refine user representations. We conduct experiments on several real-world datasets and experimental results show that the proposed SUMN outperforms state-of-the-art user representation learning methods.

The study of learning universal user representation is still in the early stage. SUMN is one of the pioneering works in this field. Although our model has shown promising performance on the tasks like user profiling and preference prediction, the generated representations are certainly not omnipotent for all user modeling applications. Developing universal user representation model that is capable of capturing diverse and evolving short-term interests is necessary for the downstream tasks like commodity recommendation. How to design a comprehensive and more accurate user modeling method is the focus of our future work.

\bibliography{aaai_2021.bib}


\end{document}